\documentclass{article}
\usepackage{spconf,amsmath,epsfig}

\usepackage{algorithm}
\usepackage{algorithmic}
\usepackage{amsmath}
\usepackage{amssymb}
\usepackage{caption}
\usepackage{subfigure}
\usepackage{diagbox}
\usepackage{multirow}

\newcommand{\eat}[1]{}

\let\OLDthebibliography\thebibliography
\renewcommand\thebibliography[1]{
  \OLDthebibliography{#1}
  \setlength{\parskip}{0pt}
  \setlength{\itemsep}{0pt plus 0.3ex}
}

\begin{document}\sloppy

% Example definitions.
% --------------------
\def\x{{\mathbf x}}
\def\L{{\cal L}}

% Title.
% ------
\title{STAR-GNN: Spatial-Temporal Video Representation for Content-based Retrieval}
%
% Single address.
% ---------------
\name{Guoping Zhao$^{1,2}$, Bingqing Zhang$^{1,2}$, Mingyu Zhang$^{1,2}$, Yaxian Li$^{1,2}$, Jiajun Liu$^{4*}$, Ji-Rong Wen$^{1,2,3}$}
\address{$^{1}$School of Information, Renmin University of China, Beijing, China\\
$^{2}$Beijing Key Laboratory of Big Data Management and Analysis Methods, Beijing, China\\
$^{3}$Gaoling School of Artificial Intelligence, Renmin University of China, Beijing, China\\
$^{4}$Data 61, CSIRO, Pullenvale, Australia}
\maketitle

\renewcommand{\thefootnote}{\fnsymbol{footnote}}
\footnotetext[1]{Corresponding author. Jiajun.liu@csiro.au}

\begin{abstract}
  We propose a video feature representation learning framework called STAR-GNN, which applies a pluggable graph neural network component on a multi-scale lattice feature graph. 
  The essence of STAR-GNN is to exploit both the temporal dynamics and spatial contents as well as visual connections between regions at different scales in the frames. It models a video with a lattice feature graph in which the nodes represent regions of different granularity, with weighted edges that represent the spatial and temporal links. The contextual nodes are aggregated simultaneously by graph neural networks with parameters trained with retrieval triplet loss. 
  In the experiments, we show that STAR-GNN effectively implements a dynamic attention mechanism on video frame sequences, resulting in the emphasis for dynamic and semantically rich content in the video, and is robust to noise and redundancies. 
  % We provide analysis to demonstrate that the way STAR-GNN constructs the multi-scale feature graph generates desirable graph properties and facilitates the training process.
  Empirical results show that STAR-GNN achieves state-of-the-art performance for Content-Based Video Retrieval.
\end{abstract}
\section{Introduction}
\label{sec:intro}

The amount of video data in video-sharing APPs such as as YouTube and TikTok has grown exponentially. 
Content-Based Video Retrieval(CBVR) has attracted tremendous attention from both the academia and the industry in the past decade due to its significant role in many video applications, such as video search, video annotation, personalized recommendation and copyright infringement detection.
\begin{figure}[htbp]
  %\centering
  %\hspace{-0.3cm}
  \includegraphics[width=0.48\textwidth]{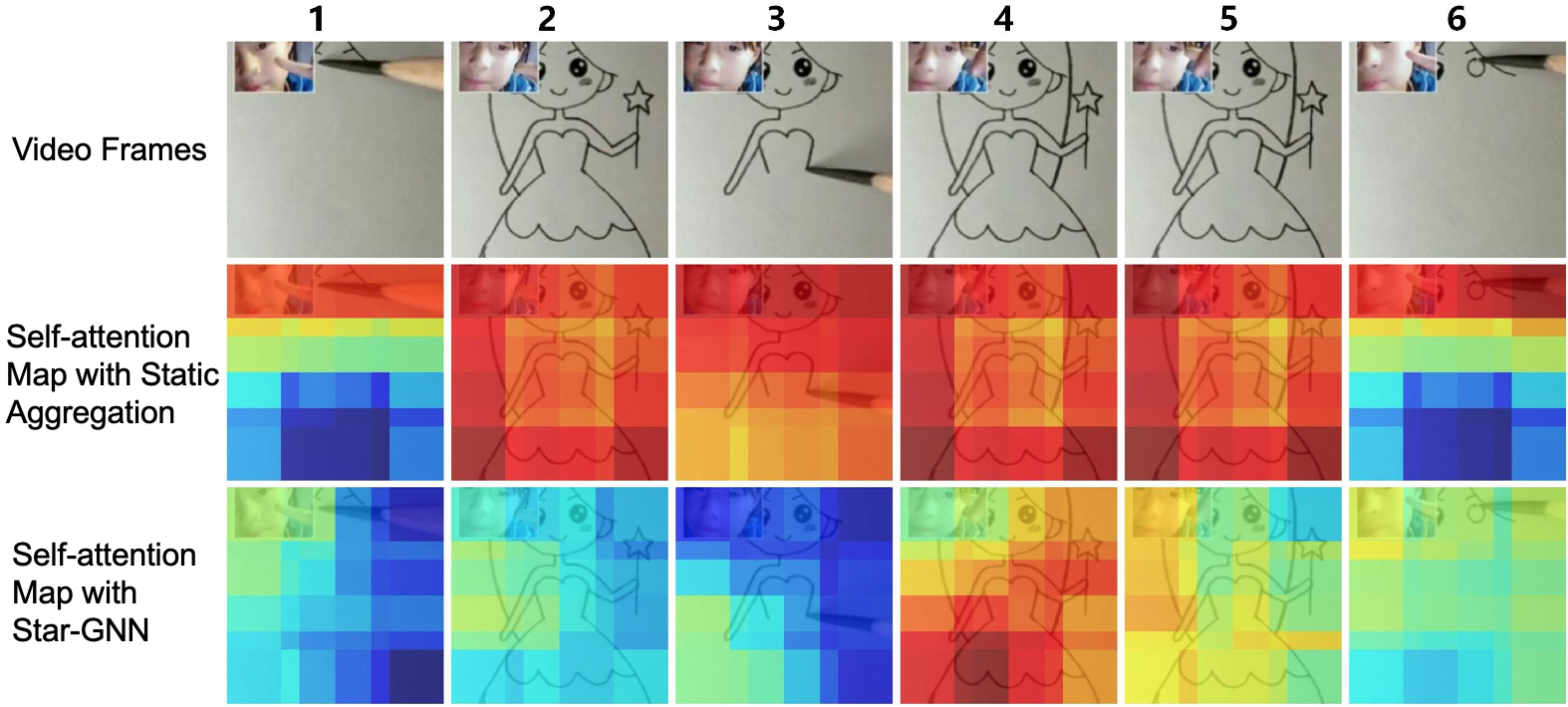} 
  \caption{STAR-GNN achieves a dynamic attention mechanism with which dynamic and semantic-rich regions in the frame sequences are given more emphasis in the final video feature representation. Over-representation of repeated elements and background is alleviated.}
  \label{fig:example}
  %\vspace{-0.5cm}
\end{figure}
The performance of CBVR depends on representativeness and discriminativeness of the video features. 
There are two common video feature representation extraction/learning methods based on their granularity: frame-level\cite{li2019w2vv++} methods and video-level\cite{kordopatis2017near} methods.
The feature representation of a video in frame-level methods is the set of frame features, and the similarity between videos is determined by the similarity of the frames features from the videos.
Such frame-level methods have major limitations. 
They regard a video as a series of static images without integrating spatial and temporal information contained in the video.
In addition, frame-level methods have a signiﬁcantly larger memory and complexity costs.
Video-level methods use a compact feature to represent the entire video.
An intuitive approach to generate video-level representations is to aggregate the frame-level CNN features into a compact global descriptor, e.g., by using global max-pooling \cite{tolias2015particular}, global average-pooling \cite{babenko2015aggregating} or LSTM\cite{gu2016supervised}. 
Moreover, the way of feature aggregation by directly pooling the frame-level features is prone to visual noise, video transformations, and content redundancy.
In order to address these current issues, we propose a spatio-temporal graph-based framework for CBVR, called Spatial-TemporAl video Representation GNN (STAR-GNN). We first split frames into interlinked regions and then represent a video via a lattice feature graph. By transforming and aggregating the spatio-temporal context of regionalized video features with pluggable GNN modules, it effectively implements a dynamic attention mechanism in the process of video feature extraction.

Figure \ref{fig:example} depicts the dynamic attention effect that STAR-GNN achieves by a visualization method named self-attention map. It shows an attention heat map between a video's final representation and its original frame regions. 
The first row shows the original video content, while the second and third rows illustrate the self-attention heat maps generated from the static aggregation and STAR-GNN, respectively.
Static aggregation directly aggregates frame regions' CNN features with a fixed regionalization scheme without modeling the spatio-temporal context. 
Evidently, for identical/near-identical frames (2, 4, 5 and 1, 6 in Figure~\ref{fig:example}), the attention heat maps are identical too. The painter's face is highly repetitive in the sequence, and as a result it is over-represented in the final video feature. Similar phenomenon is observed for the drawing in frames 2, 4 and 5. On the contrary, STAR-GNN's video representation is highly representative for semantically rich regions and dynamic contents through time. STAR-GNN represents the content of the anime figure mainly in frame 4, and then captures the changes in frame 3 as the bottom part disappears. For frames 2 and 5, which are highly repetitive, STAR-GNN lowers its attention on them. It also preserves the visual information for the painter in a more reasonable way. Across the entire sequence, it is evident that STAR-GNN always focuses on the key and dynamic visual content and overcomes the over-representation problem.

STAR-GNN is a pioneer work to use graph-based spatio-temporal aggregation and contrastive learning for content-based video retrieval, and achieves SOTA results. There are a few papers using spatial-temporal aggregation on videos \cite{NEURIPS2019_383beaea, wang2018videos, yan2020learning}, however they focus on classification or person-ReID and have limitations when applied to video retrieval as they require a set of ``known classes''. 
% For example, \cite{wang2018videos} proposed Space-Time Region Graphs for video classification, but it uses object region proposal before graph construction which causes potential problems when ``unknown'' objects to the detector are present in frames, and this renders \cite{wang2018videos} unsuitable for video retrieval where anything could potentially appear. 
STAR-GNN is not limited to any ``known classes'' and is therefore more robust to video retrieval. 
Our main contributions are two-fold:
\begin{itemize}
  \item We propose an video feature extraction framework that starts with formulating a video as a lattice graph and then harnesses the representative power of pluggable GNN modules to extract distinctive features.
  \item We conduct an extensive empirical evaluation for STAR-GNN, exploring its performance with different GNN components, and on different datasets. 
  % \item We provide in-depth analysis to show that the way STAR-GNN constructs the graph leads to desirable theoretical properties that help to overcome common drawbacks of GNNs. We also investigate how STAR-GNN effectively implements a dynamic attention mechanism (See supplementary metrial) . 
\end{itemize}

\section{Related Work}
% With the development of deep learning, CNN-based CBVR methods are quickly adopted \cite{radenovic2018fine, wu2018weighted}. Tolias et al. \cite{tolias2015particular} proposed a particular object retrieval model in spirit of Faster-RCNN \cite{ren2015faster}. In their work, a pre-trained CNN model was used for searching and re-ranking.

DNN-based CBVR can be divided into frame-level methods and video-level methods.
Frame-level methods need compute the similarity of visual content between frames.
Podlesnaya et al. \cite{podlesnaya2016deep} took advantage of the pre-trained CNN to extract features of key frames, then build complicated spatial and temporal index of frames.
Kordopatis et al. \cite{kordopatis2019visil} calculate video-to-video similarity by refined frame-to-frame similarity matrices.
Han et.al\cite{han2021video} modelled the video similarity as the mask map predicted from frame-level spatial similarity.
However, most of these frame-level methods ignore the temporal sequence information between frames, and some studies have turned to video-level features.
Zhao et al. \cite{zhao2017videowhisper} emphasized the temporal dynamics of a video, and they proposed a self-supervised recurrent neural network to learn video-level features.
Kordopatis et al. \cite{kordopatis2017near} used the deep metric learning framework to learn feature representations of videos.
Xu et al. \cite{Xu2019Self} used 3D convolutional neural networks to extract features for clips and proposed a self-supervised model by predicting frame orders. 

Graph neural networks (GNNs) have received growing attention in recent years due to their power of modelling non-linear structures in the graph topology. Among them, graph convolutional networks (GCNs) have been widely used for feature representation. GCNs can be classified into two major categories, namely spectral graph convolutional networks \cite{kipf2016semi} and spatial graph convolutional networks\cite{wu2020comprehensive,chiang2019cluster}. 
GCNs have been used to model video features in some scenarios, such as action recognition \cite{yan2018spatial}, object relation detection \cite{wang2018videos} and video person ReID \cite{yan2020learning}.
Different from this work, our STAR-GNN is the first to apply the GNNs for video retrieval task.
\begin{figure*}[htp]
  \vspace{-1cm}
      \centering
      \includegraphics[width=\textwidth]{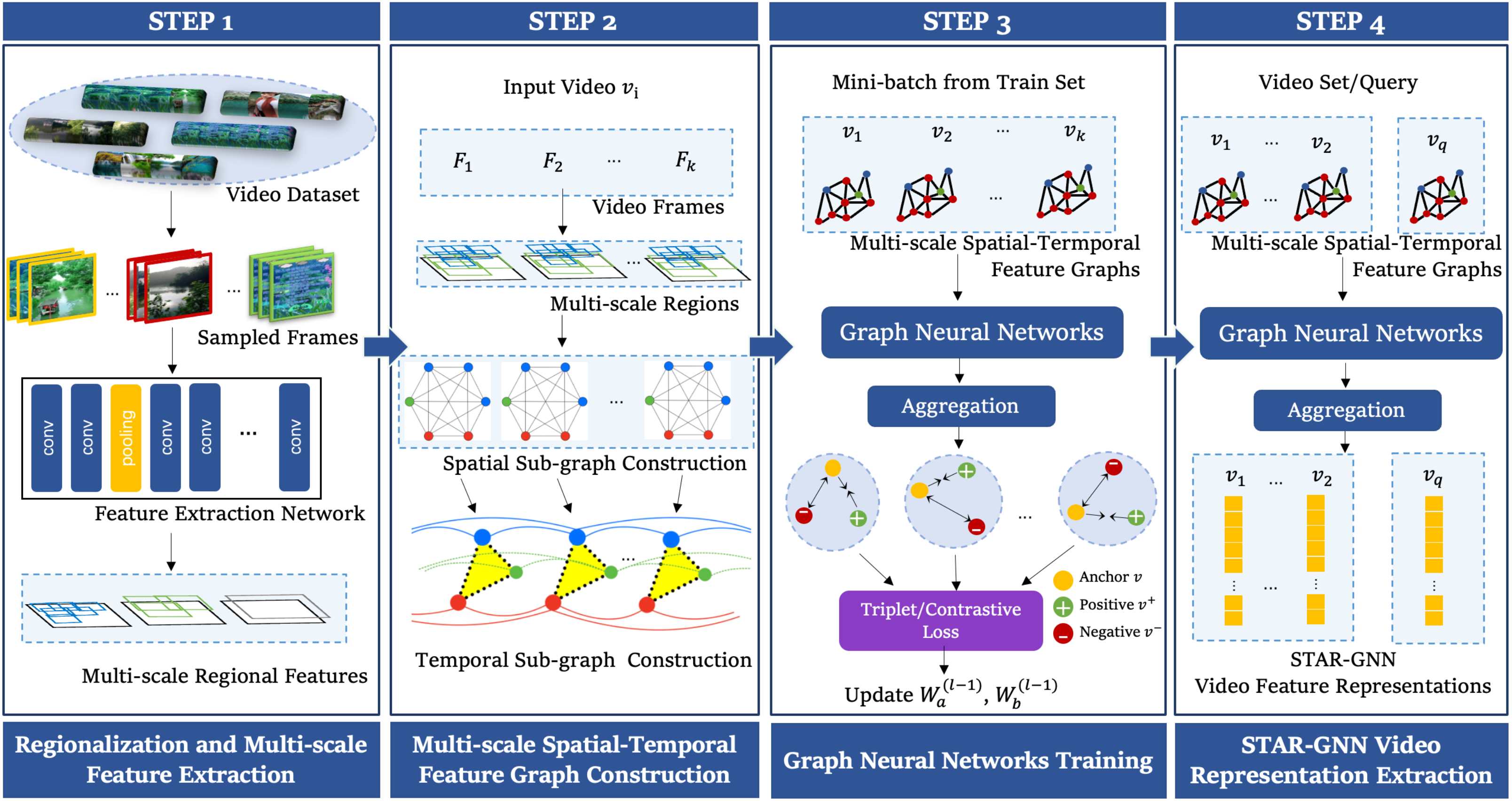} 
      \caption{Overview of the proposed STAR-GNN model.}
      \label{method}
      \vspace{-0.5cm}
  \end{figure*}

  \section{The STAR-GNN Framework}
\subsection{Overview}

We introduce the proposed STAR-GNN framework for video feature extraction. As illustrated in Figure~\ref{method}, the training process of STAR-GNN can be described in four steps.
First, for a video, we sample frames at a given rate and take the pre-trained backbone network to extract the raw frame feature map. Multi-scale regional features are then generated from the frame feature map.
Second, each video is transformed into a regional graph with spatial-temporal links.
Third, we train the graph neural network in the STAR-GNN framework with retrieval triplet loss or contrastive loss.
Fourth, each regional graph is processed with the trained STAR-GNN using shared parameters. The feature for a video is represented as the average of all regional node representations in its regional graph.
The rest of this section provides further details.

\subsection{Spatial-Temporal Graph Construction}
\label{ST-Graph Construction}

%yaxian
A video consists of frames with a temporal order, and each frame contains semantic contents appearing in various locations. A video is therefore a collection of inter-related temporal-spatial information in three dimensions. To integrate these aspects into a unified feature space, we construct an undirected/weighted graph $G(V, E)$ for each video, to model both the spatial and temporal domains.

First, we divide a single frame into multi-scale regions to model the local visual coherence and enable the modeling of regional relationships through time. Considering that the regions of interest among different frames may vary in size, we use $s$ different scales. These scales are applied on frame feature maps extracted by the backbone network to obtain region-level features with corresponding strides. Consequently, each frame is split into several regions, with each region corresponding to a node in the regional graph. And a frame $F_i$ will have $s$ types of regions which are represented by $R^{ki}_{j}$, in which $j$ stands for the region index and $k\in\{1,2,...,s\}$ is the scale index. See Figure~\ref{method} STEP 1. 

Second, to take advantage of a video's local temporal dynamics, we inter-connect all regional nodes at the same spatial location across all frames in a video to form a complete temporal sub-graph. The graph construction strategy of STAR-GNN is formally defined as (STEP 2 in Figure~\ref{method}):
\begin{enumerate}
    \item \emph{Node formulation}: Each region $R^{ki}_{j}$ ($k\in\{1,2,...,s\}$) is represented as a node in graph $G$. 
And the edge weight $w_{ij}$ between node $n_i$ and node $n_j$ is the cosine similarity of corresponding region features.
    \item \emph{Spatial Connectivity}: Every pair of regions belonging to the same frame are connected, i.e. for a fixed $i=\hat{i}$, the sub-graph defined by the node set $\{R^{k\hat{i}}_{j}\}$ and the edges among them is a complete graph.
    \item \emph{Temporal Connectivity}: Regions of the same scale and the same location are connected through different frames. Given $j=\hat{j}$, the sub-graph defined by the node set $\{R^{\hat{k}i}_{\hat{j}}\}$ and the edges among them is a complete graph.
\end{enumerate}

The spatio-temporal graph constructed with the procedure described above has a few desirable properties for a GCN. Details are provided in the Analysis part.

\subsection{GNN for Spatio-temporal Feature Extraction}
\label{cluster-GCN optimization}
We consider an undirected graph $G(V,E)$. When applied to feature representation learning for a graph, graph neural networks are designed to learn a transition function $f$ as well as an output function $g$. Such that:
\begin{equation}
H = f(H,X),
\end{equation}
\begin{equation}
O = g(H,X_N),
\end{equation}
where $X$ is all features, $H$ is the graph's hidden states, $X_N$  denotes all the node features and $O$ denotes the final feature representation for all the graph nodes. \eat{In order to gain latent feature representation more effectively, the graph convolution operation replaces the local transition function $f$ with a filter $g_{\theta} = \mathrm{diag}(\theta )$ parameterized by $\theta \in \mathbb{R}^{N}$. Finally, it is defined as $g_{\theta} \star x = U g_{\theta} U^{T} x$.  }

In this STAR-GNN frame work, instead of outputting all node representations, we use a global aggregation function to compress all these learned features into a single feature vector to represent the entire graph. That is, the graph neural networks in the STAR-GNN framework can be written as:
\begin{equation}
o = AGGREGATE(G_\Theta \star X).
\end{equation}
Here, $G_\Theta$ is the GNN layer-wise operator and $AGGREGATE$ is the aggregate function which takes in a feature matrix for all the graph nodes and outputs a single feature vector. 

In particular, the GNN layer-wise operator here can be replaced by any specific GNN implementation, such as the vanilla GCN layer  \cite{kipf2016semi}  $H^{(l+1)} = \sigma (\hat{A} H^{(l)} W^{(l)})$, cluster GCN layer \cite{chiang2019cluster} $H^{(l)} = \sigma (\hat{A} H^{(l-1)} W_{a}^{(l-1)} + H^{(l-1)} W_{b}^{(l-1)})$ and so on. Here, $\hat{A}$ is a renormalized 
adjacency matrix, $W$ denotes a 
learnable parameters. Correspondingly, $AGGREGATE$ function can be replaced by different strategies such mean aggregation, max pooling aggregation and so on.

In order to train STAR-GNN, we use the GCN mini-batch training strategy. But instead of mini-batches of nodes from the same graph as in most of the literature \cite{chiang2019cluster,hamilton2017inductive}, we spilt the training video set into mini-batches with $B$ videos in each, and use the set of spatio-temporal feature graphs that correspond to the videos as the input signal to STAR-GNN. We then use triplet loss to update the model’s parameters at each iteration on a mini-batch, which is defined as:
\begin{equation}
\mathcal{J} = \sum_{i=1}^{B} \mathcal{L}(AG^{(i)},PG^{(i)},NG^{(i)}), 
\end{equation}
$\mathcal{L}$ can be a triplet loss or contrastive loss, $AG^{(i)}$ denotes the output graph features for an anchor video sample, $PG^{(i)}$ denotes that for a positive retrieval sample and $NG^{(i)}$ is that of a negative sample. Here we regard a labelled positive sample in the dataset for a given query $AG^{(i)}$ as $PG^{(i)}$, and a random sample outside the labelled positive sample sets as $NG^{(i)}$. The gradient is then updated for each mini-batch iteration:
\begin{equation}
    \frac{1}{B} \sum_{i \in \mathcal{B} } \bigtriangledown \mathcal{L}(AG^{(i)},PG^{(i)},NG^{(i)}).
\end{equation}

% Next, we analyze the graph topology in the STAR-GNN model and discuss its interesting properties and how they are beneficial for the overcoming of over-smoothing, over-squashing and under-reaching.

\section{Experiments}
\subsection{Experimental Settings}
\label{subsec:expset}
\noindent
{\bf Datasets.} We verify the effectiveness of STAR-GNN and compare its performance with state-of-the-art methods on the widely-used VCDB \cite{jiang2014vcdb} and the more recent SVD \cite{jiang2019svd} datasets.
The VCDB dataset contains 100,528 videos, with 528 queries and 6,139 manually annotated as positive retrieval samples. 
It was initially used for partial copy detection in videos, and then also widely used to evaluate the performance of video retrieval models.
The SVD dataset is recent introduced for large-scale short video retrieval, which contains 562,013 short videos from Douyin (TikTok China).
It contains 1,206 queries and 34,020 labeled videos, of which 10,211 are positive retrieval samples and 26,927 are negative.
To make a fair comparison, we follow the setting in \cite{jiang2019svd}: we use $1,000$ labeled queries for training and the other $206$ labeled queries for testing. We then take the model trained on SVD-train to perform further tests with VCDB. 
% For DML \cite{kordopatis2017near} we use the same setting.

\noindent
{\bf Pluggable GNN Methods.} We use different GNN methods to verify the stability of STAR-GNN. Initially we use vanilla GCN \cite{kipf2016semi}, which is one of the most classic GNN models. In addition, we also estimate the effects of SGCN \cite{DBLP:conf/icml/WuSZFYW19} as well as Cluster-GCN. \cite{chiang2019cluster}.

\noindent
{\bf Evaluation Metrics.} We measure the retrieval performance with mean Average Precision (mAP), which is a standard evaluation protocol in video retrieval.

\noindent
{\bf Implementation Details.} For frame features, we utilize the feature maps output from the last convolutional layer of the backbone network, which is VGG16 pre-trained on Imagenet in our experiments.
% mingyu
A raw video's smaller edge is resized into 256, then center-cropped into 224 $\times$ 224, and is extracted with the feature map of 7 $\times$ 7 $\times$ 512.
We then use sliding windows of three different scales, namely $3 \times 3$, $4 \times 4$  and $7 \times 7$, to obtain 9, 4 and 1 regional features respectively. Consequently, each frame is split into 14 regions. Then, we apply the global max-pooling operation on the regional feature. After extracting node features from STAR-GNN, we use zero-mean and \L2-normalization, to generate the final video-level features. And We set $S=1$ for STAR-GNN's graph neural network and set the activation function to ReLU. For training, we set the margin $\lambda$ = 0.5 in the triplet loss function, and train the model with batch size of 128.
We used online hard triplet mining strategy. For each anchor, we select the hardest negative (smallest distance to anchor) in the batch.

\begin{table*}[h]
\vspace{-0.5cm}
    \centering
    \caption{Comparison of Retrieval mAP with State-of-the-Art CBVR methods (higher is better).}
    \label{tab:stoa}
    \begin{tabular}{l|lllll|llll}
    \hline
    \multirow{2}{*}{\diagbox{Method}{Distraction}}                              & \multicolumn{5}{c|}{SVD}                                                                                                         & \multicolumn{4}{c}{VCDB}                                                                             \\  \cline{2-10} 
                & \multicolumn{1}{r}{0} & \multicolumn{1}{r}{10K} & \multicolumn{1}{r}{50K} & \multicolumn{1}{r}{100K} & \multicolumn{1}{r|}{520K} & \multicolumn{1}{r}{0} & \multicolumn{1}{r}{10K} & \multicolumn{1}{r}{50K} & \multicolumn{1}{r}{100K} \\ \hline

    \hline
    MAC\cite{tolias2015particular}                     & 0.8974                & 0.8031                  & 0.7787                  & 0.7725                   & 0.7499                    & 0.7307                & 0.6268                  & 0.5405                  & 0.5068                   \\
    R-MAC \cite{tolias2015particular}                   & 0.8981                & 0.8163                  & 0.7918                  & 0.7852                   & 0.7642                    & 0.7362                & 0.6395                  & 0.5524                  & 0.5170                   \\
    SPOC  \cite{babenko2015aggregating}                  & 0.8845                & 0.8002                  & 0.7745                  & 0.7643                   & 0.7271                    & 0.6849                & 0.5786                  & 0.5003                  & 0.4667                   \\ 
    3D-shufflenetv2\cite{kopuklu2019resource}         & 0.4244                & 0.0019                  & 0.0004                  & 0.0002                   & 0.0000                    & 0.1258                & 0.0724                  & 0.0697                  & 0.0693                   \\ 
    DML\cite{kordopatis2017near}                     & 0.9063                & 0.8287                  & 0.8074                  & 0.8014                   & 0.7706                    & 0.7681                & 0.6670                  & 0.5760                  & 0.5373                   \\
    3D-shufflenetv2(Supervised)       & 0.6532	& 0.1413	& 0.1108	& 0.1023	& 0.0816     & 0.2102	& 0.1328	& 0.1221	& 0.1204                  \\ 
    RGB\cite{wang2021attention}                     & 0.9140    & -    & -      & -     & -     & -     & -     & -     & -    \\ 
    \hline
    \hline
    STAR-GNN(CL+Cluster-GCN) & 0.9278	& \textbf{0.8812}	& \textbf{0.8597}	& \textbf{0.8528}	& \textbf{0.8234}	& 0.7561	&0.6675	& 0.5888	& 0.5541         \\
    STAR-GNN(TL+Cluster-GCN) & \textbf{0.9329}       & 0.8782         & 0.8563        & 0.8462          & 0.8110           & 0.7954       & 0.714          & 0.6356         & 0.5997          \\ 
    STAR-GNN (TL+vanilla GCN) & 0.9216 & 0.7965 & 0.7965 & 0.7965 & 0.7965 & 0.7742 & 0.6665 & 0.5807 & 0.5447 \\
    STAR-GNN (TL+SGCN) & 0.9321 & 0.8704 & 0.8522 & 0.8416 & 0.8054 & \textbf{0.8025} & \textbf{0.7201} & \textbf{0.6424} & \textbf{0.6077} \\
    %Performance Gain    &  2.94\% & 5.97\% &  6.05\% &  5.59\% &  5.24\% &  3.55\% &  7.05\% &  10.35\%  &  11.61\% \\ \hline
\hline
    
    \end{tabular}
    \vspace{-1em}
\end{table*}

\subsection{Comparative Results}
In the experiments, we compare the proposed method with six state-of-the-art content-based video retrieval methods, including MAC\cite{tolias2015particular}, RMAC\cite{tolias2015particular}, SPOC\cite{babenko2015aggregating}, 3D-shuffenetv2\cite{kopuklu2019resource}, DML\cite{kordopatis2017near},  3D-shuffenetv2(Supervised)\cite{kopuklu2019resource} and RGB\cite{wang2021attention}.
Among them, MAC, RMAC, and SPOC used VGG pre-trained on ImageNet as feature extractor, and 3D-shuffenetV2 used a 3D convolutional network pre-trained on video classification dataset Kinetics-600\cite{kay2017kinetics}.
For DML and 3D-shuffenetV2(Supervised), we use the same training set with STAR-GNN.
RGB is a attention-based deep metric learning model, which we use the data from the source paper.
For STAR-GNN, CL denote Contrastive Loss and TL denote Triplet Loss. 
Comparative results for retrieval mAP on SVD and VCDB datasets are shown in Table~\ref{tab:stoa}.
The top half of Table~\ref{tab:stoa} includes the results for some previous methods while the lower half shows the methods using STAR-GNN framework. 
Additional \emph{Distraction} stands for the number of distractor videos, which are randomly selected, unlabeled videos (apart from the positive samples and negative samples labelled) in test datasets. 
They drastically increase the volume of the candidate set and raise the difficulty significantly for retrieval. For each configuration, we use bold text to mark the best performing results.

From the Table~\ref{tab:stoa}, we can see that STAR-GNN outperform all other baselines under all \emph{distraction} settings.
On SVD, our method gives mAP=0.9329 without distractor, which is a 2.67\% relative improvement compared to the current state-of-the-art supervised CBVR method, i.e., DML\cite{kordopatis2017near}, and a 5.28\% relative improvement when the number of distractor videos is 520,000.
A similar trend was reported on VCDB dataset, STAR-GNN outperforms other baselines with a consistent margin (+0.0273~+0.0624) on VCDB.
Importantly, with the size of the distraction set increasing, the mAP of all methods will decrease, but STAR-GNN decreases the least, which means STAR-GNN is significantly more stable to the feature representation of the videos.
On SVD, with an increasing additional distraction set from 0 to 520K, the mAP  of our proposed method are 92.78\%, 88.12\%, 85.97\%, 85.28\%, 82.34\%, respectively.
On VCDB, the mAP  of our proposed method are 80.25\%, 88.12\%, 72.01\%, 64.24\%, 60.77\%, respectively.
All reported results show that STAR-GNN has signiﬁcant beneﬁt to content-based video retrieval.

% \begin{table}[htp]
% \centering
% \caption{Comparison of feature size and run time.}
% \begin{tabular}{|c|r|r|}
% \hline
% Method & Feature Dim & $T_{process}$\\ \hline
% MAC                         & 512                                    & 0.0572                                                            \\\hline
% 3D-shufflflenetv2           & 1024                                   & 0.1359                                                             \\\hline
% DML                         & 500                                    & 0.0668                                                             \\\hline
% STAR-GNN                     & 512                                    & 0.1755                                                             \\ \hline
% \end{tabular}
% \label{tab:time_space}
% \vspace{-0.5cm}
% \end{table}

\subsection{Ablation studies}
\subsubsection{Effect of Network Structure Settings}
The Table~\ref{tab:stoa} shows the evaluation results of the different components of our proposed method.
The best results are achieved by the STAR-GNN(Contrastive Loss+Cluster-GCN) on SVD and STAR-GNN(Triplet Loss+SGCN) on VCDB.
The results reﬂect that Cluster-GCN and SGCN work better than vanilla GCN.
The impact of the loss function is not certain. 
The contrastive loss performance is better on the SVD dataset, while on the VCDB, the triplet loss has an advantage.

\subsubsection{Effect of Region Size}
We use the multi-scale regions on CNN feature map to construct the spatio-temporal feature graph.
The region size affects the number of areas, so we evaluate the effect region size.
Table \ref{tab:regionSize} shows the experimental results of different region sizes. With single-scale regions, as the size decreases, retrieval performance improves. However, as we utilise combined, multi-scale regions, the method exhibits a substantial performance gain.

\begin{table}[htbp]
\centering
\caption{mAP for Different Region Size}
\label{tab:regionSize}
\begin{tabular}{|c|c|c|c|}
\hline
Region Size & Number of Regions & SVD       & VCDB   \\ \hline
7x7         & 1 (1×1)           & 0.6937    & 0.4552 \\ \hline
4x4         & 4 (2×2)           & 0.7648    & 0.5002 \\ \hline
3x3         & 9 (3×3)           & 0.7858    & 0.5246 \\ \hline
Multi-scale & 14(1+4+9)         & 0.8234    & 0.6077 \\ \hline
\end{tabular}
\vspace{-1em}
\end{table}

\subsubsection{Effect of Edge Weights}
We also evaluated whether to use the similarity between regions as the weight of edges when constructing the spatio-temporal feature graph.
Table \ref{tab:weight} verifies that by formulating the spatio-temporal feature graph into a weighted graph, STAR-GNN is able to better reflect each video's spatio-temporal characteristics through its graph topology.
It should be noted that the weighted graph can be further improved performance, but the weighted graph needs to calculate the similarity between connected node, which causes huge computational cost.
Therefore, this part of the experiment is only carried out on SVD-10K, which is a subset of SVD.
\begin{table}[htbp]
\centering
\caption{Performance Comparison (in mAP) with Weighted \& Unweighted Graphs}
\label{tab:weight}
\begin{tabular}{|c|c| c|}
\hline
            & STAR-GNN      & STAR-GNN (Unweighted)          \\ \hline
SVD-10K     & 0.9112        & 0.8812 \\ \hline
\end{tabular}
\vspace{-0.5cm}
\end{table}

\subsubsection{Effect of Training Dataset Size}
To evaluate the the impact of the training set size on the experimental results, Table \ref{tab:ratio} shows the mAP with different training dataset ratios. 
The mAP on both datasets improves as the ratio of the train set increases. For example, on SVD as we increase the train ratio from 10\% to 100\%, the mAP also climbs from 0.7749 to 0.8110. Similar results are observed on VCDB too.
This shows that as the training set increases, the retrieval performance of STAR-GNN will improve.
It is worth noting that STAR-GNN achieves competitive performance with only a small amount of labeled training data (even 10\%), and is able to learn from more labelled samples.

\begin{table}[htbp]
    \centering
    \caption{mAP for Different Train Ratio}
    \label{tab:ratio}
    \begin{tabular}{|c|c|c|c|c|c|}
        \hline
        Ratio   &10\%       &20\%       &50\%   &70\%   &100\%   \\ \hline
        SVD     &0.7749     &0.7945     &0.7981 &0.7936 &0.8110  \\ \hline
        VCDB    &0.5321     &0.5704     &0.5603 &0.5880 &0.5997  \\ \hline
    \end{tabular}
    \vspace{-0.5cm}
\end{table}

% \vspace{-0.2cm}
\section{Conclusion}
We propose a novel method called STAR-GNN as a video feature extractor for CBVR. STAR-GNN employs graph neural networks to perform spatial-temporary context aggregation so that the video feature representation is sensitive to the dynamic and semantic-rich part. We show that it effectively implements a dynamic attention mechanism for CBVR, hence is robust to the perturbation and over-representation introduced by back banners, logos, subtitles, overlays, or identical frames, etc, in the video. We also show the spatio-temporal feature graph that STAR-GNN constructs have a few properties that are beneficial to the quality of GNN-generated features. The method achieves SOTA retrieval performance on two public datasets and exhibits strong evidence for its capability of preserving dynamic and semantically meaningful contents in its feature representation. In the future, we will extend our work into video recommendation and caption.

\bibliographystyle{IEEEbib}
\bibliography{icme2022template}

\end{document}